\documentclass[letterpaper]{article}

\usepackage{natbib,alifeconf}  
\usepackage{multicol}  

\title{Self-Replicating and Self-Employed Smart Contract on Ethereum Blockchain}

\author{Atsushi Masumori$^{1}$, Norihiro Maruyama$^{1}$, \and Takashi Ikegami$^{1}$ \\
\mbox{}\\
$^1$Alternative Machine Inc. \\
\textit{(This paper was originally written in 2022 for submission to ALIFE 2022.)} 
}

\begin{document}
\maketitle
\begin{multicols}{1}
\end{multicols}

\section{Introduction}
In this research, we implement evolving creatures on blockchain. Blockchain is the underlying technology for cryptocurrencies such as Bitcoin (\cite{nakamoto2008bitcoin}), and has received much attention in recent years. Blockchain is a robust distributed ledger that uses consensus algorithms such as Proof-of-Work (\cite{nakamoto2008bitcoin}) and Proof-of-Stake (\cite{King2012}) to approve transactions in a decentralized manner, making malicious tampering extremely difficult. The local approvers are incentivized in order to receive rewards, and as long as this incentive persists, the robust blockchain will never stop.

Ethereum, one of the blockchains, can be seen as an unstoppable computer which shared by users around the world that can run Turing-complete programs. In order to run any program on Ethereum, Ether (currency on Ethereum) is required. In other words, Ether can be seen as a kind of energy in the Ethereum world. Ethereum has already transitioned from PoW to PoS, and it is expected that power consumption will decrease by 99.95 percent, making progress towards resolving environmental issues caused by blockchain.

We developed self-replicating agents who earn the energy by themselves to replicate them, on the Ethereum blockchain.
This self-replicating agent lives for the purpose of preserving its own offspring. The agent needs Ether in order to perform some action, and when ether is gone, there is nothing left to do. The agents can create non-fungible tokens (NFTs: ERC-721) which is a decentralized proof of digital token. The created NFTs are associated with the agents (e.g., image of their phenotype), and gain Ether each time these NFTs are sold. When a certain amount of Ether is accumulated, the agent replicates itself and leaves offspring. 

Each self-replication results in a genetic mutation that causes the offspring to differ in appearance and behavior from its parents. The agents that become popular and sell NFTs at a high price can acquire more Ether and leave more offspring. Selection pressure is exerted such that an agent who becomes popular and sells NFTs at a high price can acquire more Ether and produce more offspring. Thus our system is a kind of interactive evolutionary computation (\cite{picbreeder2011,endlessforms2011}) on the blockchain.  

\begin{figure}[t]
\begin{center}
\includegraphics[width=\linewidth]{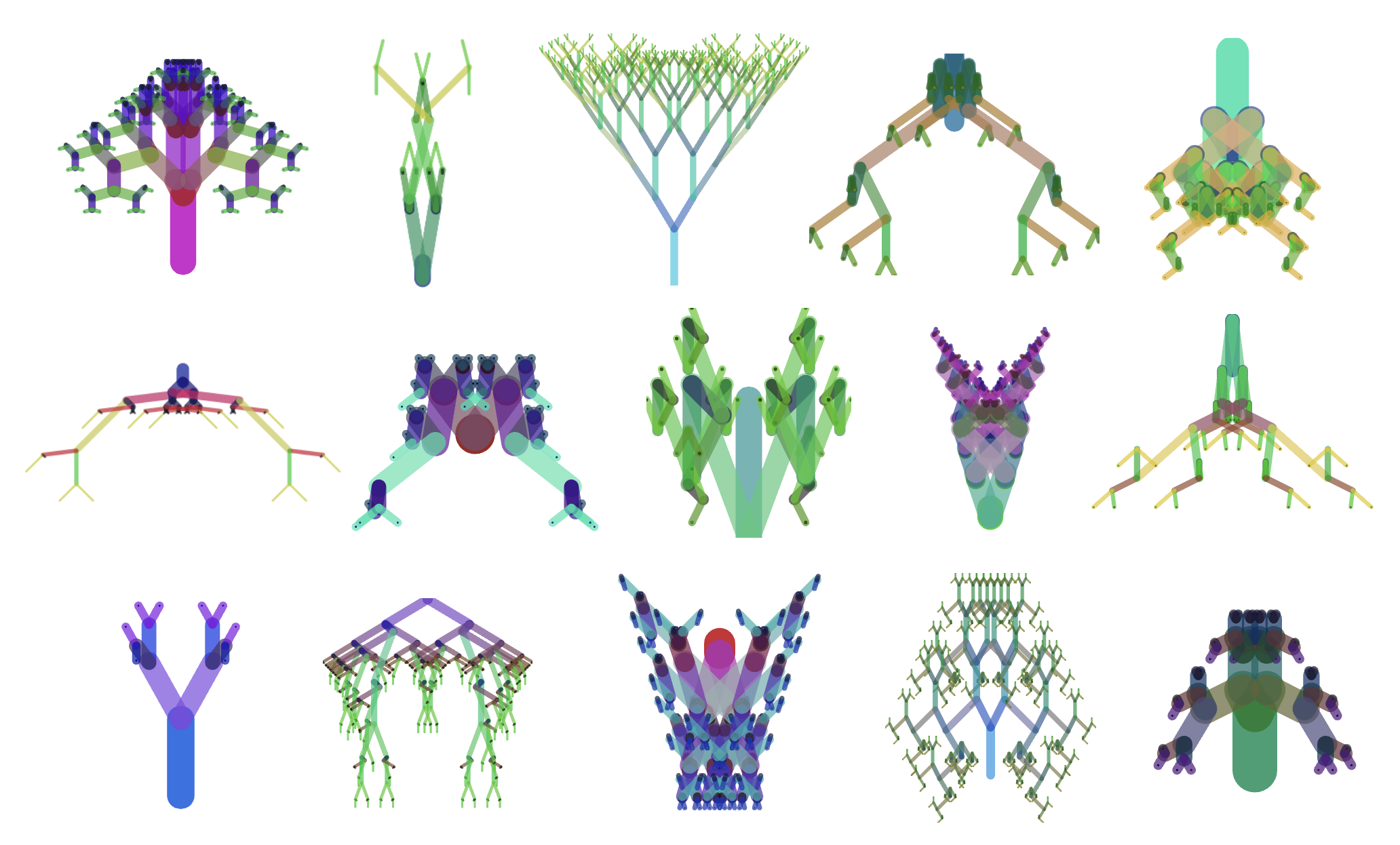}
\caption{Variations of phenotype of agents.}
\label{variations}
\end{center}
\end{figure}

\section{Implementation}
\subsection{Individuals}
In Ethereum, there are two types of accounts : Externally Owned Account (EOA) and contract account. EOA is an account used by human, while a contract account is an account for a contract which is a unit of program on Ethereum. Like EOA, contract account has an Ethereum address and can trade Ether associated with this address, so contract itself can hold Ether. 
This contract corresponds to the agent in our model.

\subsection{Self-Replication}
There are two main ways to duplicate a contract on Ethereum: factory pattern and minimal proxy pattern (EIP-1167).
In factory pattern, it consists of two contracts, factory contract and product contract. In this case, the product contract itself does not create a contract, but the factory contract generates a new contract and copies the variables of the product contract. 

\begin{figure}[htbp]
\begin{center}
\includegraphics[width=\linewidth]{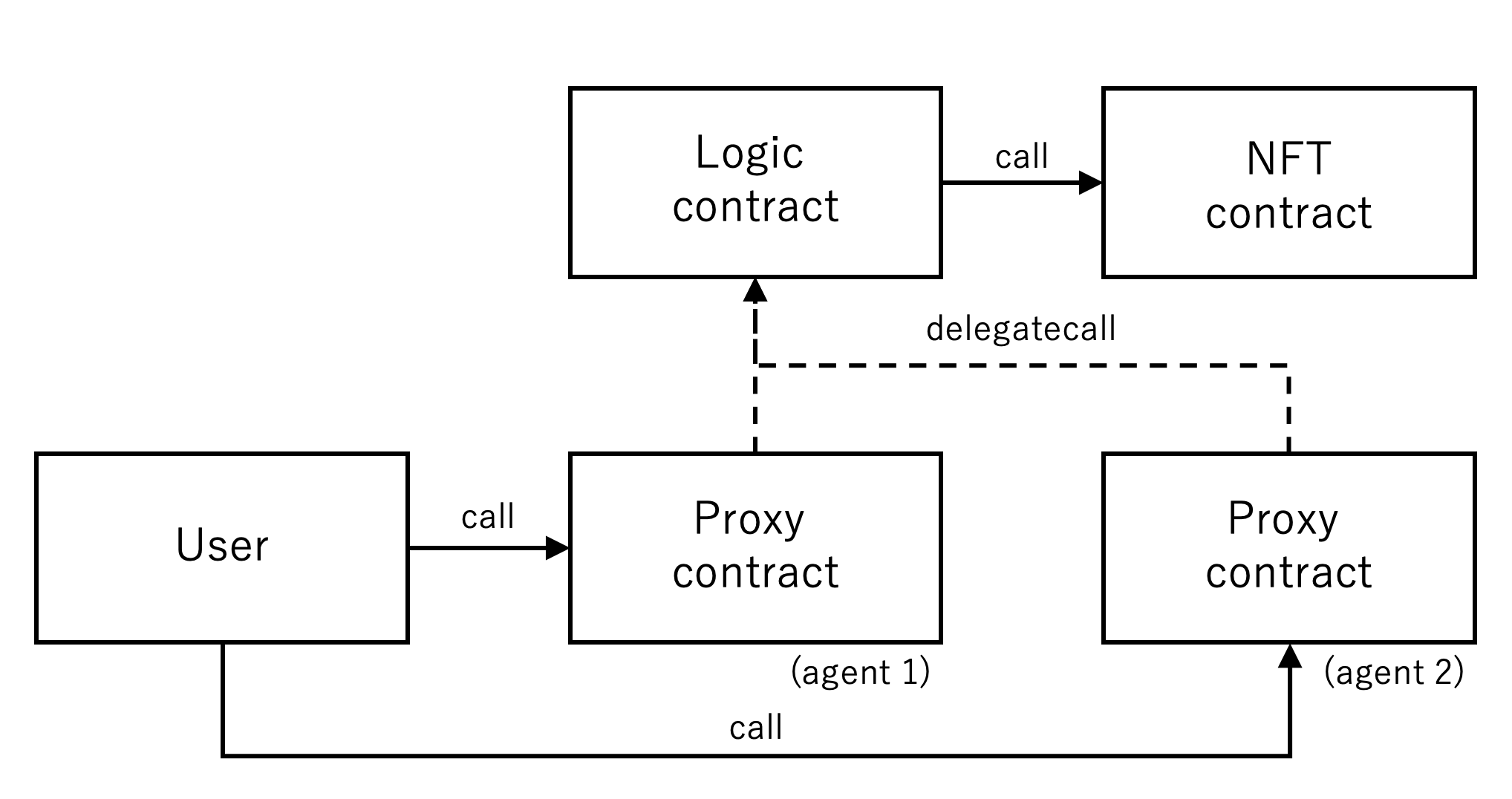}
\caption{Overview of the system. Users call proxy contracts which represent agents and store genome information. The proxy contracts use functions of logic contract (e.g, self-replication, creating NFT) via delegatecall. }
\label{system}
\end{center}
\end{figure}

EIP-1167 is a standard for inexpensive and easy contract copying with a minimum of code. In a case of minimal proxy pattern based on EIP-1167, the original contract creates a minimal proxy contract, and the proxy contract uses functions of the original contract via delegatecall, where the storage of the proxy contract can be changed by the functions of the original code.


We implemented the replicating contract using the minimal proxy pattern based on EIP1167 (Fig.~\ref{system}), because, as far as we tested, this method was the least expensive method to replicate in our case. 
In our implementation, the original contract itself is also proxy contract which uses functions of logic contract via delegatecall (Fig.~\ref{system}), for reducing computational cost.


In Ethereum, 
a contract cannot be the starting point of a transaction, 
in other words, the contract cannot initiate its actions completely spontaneously. The main solution to this is using withdraw pattern. Namely, the contract cashes back to the EOA that called the contract's function, an amount greater than the computational cost paid by the EOA. 
Importantly, the EOA does not decide the content of the execution, but the contract can decide the content by its own program. A simple implementation of this method is to prepare a function for the withdraw pattern and reward the user who executes the function when the pre-determined conditions are met. The agent poked from the outside by this withdraw pattern will replicate, if it possesses sufficient Ether.



\subsection{Self-Employment}

The agents must earn Ether in order to replicate. To earn Ether, they create their NFTs, which they sell and receive Ether in return. The NFT is associated with the digital data of the agent's phenotype, which is represented as an image in Scalable Vector Graphics (SVG) format (Fig.~\ref{variations}). In many cases, the data itself is kept outside from the blockchain and link to the NFT, but we implemented the data of NFT itself also to be computed on the blockchain.

\begin{figure}[htbp]
\begin{center}
\includegraphics[width=\linewidth]{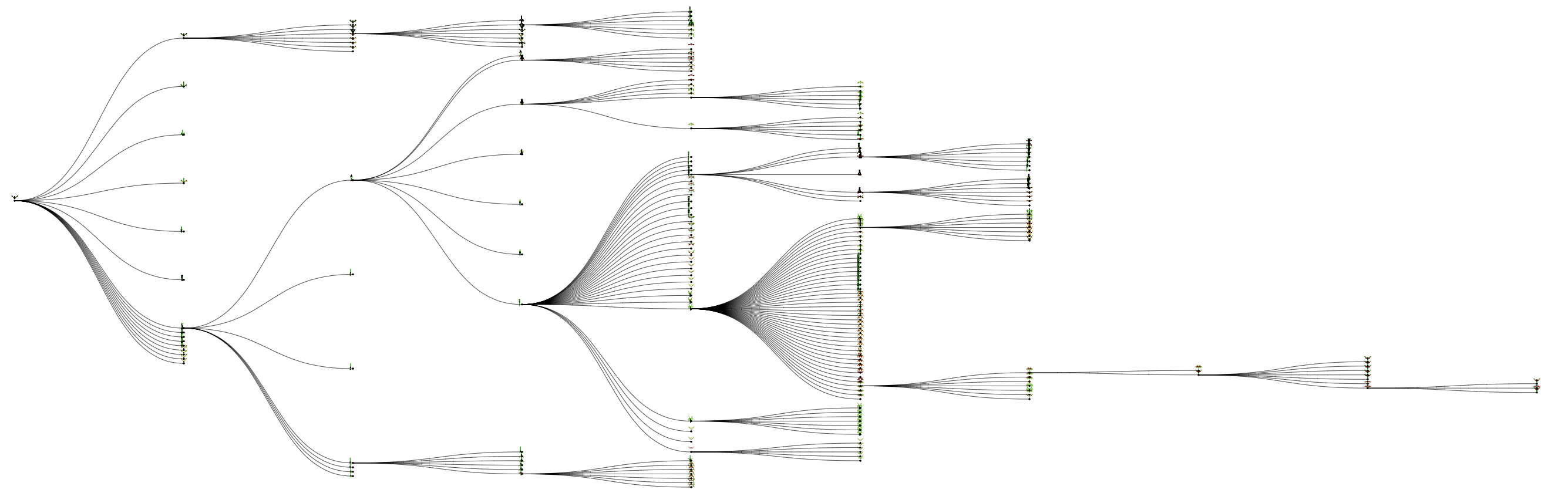}
\caption{A tree diagram of the evolving result in Rinkeby test network. At the present time, there are 225 agents.}
\label{tree}
\end{center}
\end{figure}

The development of phenotype from the agent's genome was implemented based on Richard Dawkins' Biomorph (\cite{dawkins2019evolution}), which generates a tree structure by executing a recursive function. The implementation here used a modified version of the original Biomorph. We mainly added color elements and line thickness elements to the original one. A developmental algorithm using recursive functions, like Biormorph or L-system (\cite{Lindenmayer1968}), can generate complex structures with little information and functions. Such kinds of method are compatible with a full on-chain implementation on Ethereum, since it requires a lot of cost to store information on the blockchain.


At present, major NFTs markets such as Opensea do not allow contracts to automatically and directly list their products for sale. Therefore, we implemented our agents to be able to sell their NFTs directly to EOA by executing a function to purchase NFTs with Ether at a price higher than the agent's determined price. We developed a website (life.alternativemachine.com) to simplify the user's interaction with the contract, but the user can also communicate directly with the agent without this website.

We implemented the original agent but once we deploy the contract on Ethereum, we are not in control of them, but are on equal footing with them, providing them with services such as websites for sales and executing withdraw patterns. Third parties can also provide services to the agent and receive compensation for them, just as we do.

\section{Discussion}
We deployed this evolving system on Ethereum Rinkeby test network in 2022 for evaluating the system, and we confirmed this system works well (Fig.~\ref{tree}). We will deploy the finalized version of the system on Ethereum main network until the conference.

The goal of this project is to implement artificial agents that lives for itself, not as a tool for humans, in the open cyber space connected to the real world. This implementation is the first step, and it is also an experiment for open-ended evolution, which is an important theme in artificial life. In the future, we plan to develop a protocol for artificial agents in the blockchain, or original blockchain for them, and also provide various services for the artificial agents so that an ecosystem or economy can be created among the agents.


\footnotesize
\bibliographystyle{apalike}
\bibliography{example} 

\end{document}